\documentclass[preprint,nopreprintline]{elsarticle}

\usepackage[a4paper,textwidth=440pt,textheight=640pt,centering,headheight=14pt]{geometry}

\usepackage[T1]{fontenc}
\usepackage{lmodern}
\usepackage{amsmath}
\usepackage{amssymb}
\usepackage{bm}
\usepackage{graphicx}
\graphicspath{{./}}
\usepackage{booktabs}
\usepackage{array}
\usepackage{xcolor}
\usepackage{caption}
\usepackage{placeins}
\usepackage{needspace}
\usepackage[hyphens]{url}
\usepackage{microtype}
\usepackage[hidelinks]{hyperref}
\renewcommand{\appendixname}{Appendix}

\journal{Computers and Geotechnics}

\newcommand{\paperfigure}[3]{%
  \IfFileExists{#1.pdf}%
    {\includegraphics[width=#2]{#1.pdf}}%
    {\fbox{\parbox[c][#3][c]{\dimexpr#2-2\fboxsep-2\fboxrule\relax}%
      {\centering\ttfamily\detokenize{#1.pdf}\\[3pt]\normalfont\small not adopted yet}}}%
}

\begin{document}

\begin{frontmatter}

\title{Subsurface schematisation from sparse CPT data\texorpdfstring{\\}{ }using the tabular foundation model TabPFN}

\author[label1,label2]{Taiga Saito\corref{cor1}}
\ead{taiga.saito.r3@dc.tohoku.ac.jp}
\cortext[cor1]{Corresponding author}

\author[label2]{Fabian Campos Montero}
\author[label2]{Bruno Zuada Coelho}
\author[label2]{Ana Martins Teixeira}
\author[label3,label4]{Stephen~Wu}
\author[label2]{Timo Schweckendiek}
\author[label1]{Yu Otake}

\affiliation[label1]{
    organization={Department of Civil and Environmental Engineering, Tohoku University},
    city={Sendai},
    country={Japan}
}

\affiliation[label2]{
    organization={Safe and Resilient Infrastructure, Deltares},
    city={Delft},
    country={the Netherlands}
}

\affiliation[label3]{
    organization={Research Organization of Information and Systems, The Institute of Statistical Mathematics},
    city={Tachikawa},
    country={Japan}
}

\affiliation[label4]{
    organization={The Graduate University for Advanced Studies, Department of Statistical Science},
    city={Tachikawa},
    country={Japan}
}

\begin{abstract}
Subsurface schematisation supports geotechnical decisions but remains challenging because sparse site-investigation data leave substantial gaps in information about ground conditions.
This study explores TabPFN, a foundation model pretrained on general synthetic tabular data, for this task.
Using only grid coordinates and soil behaviour type index observations from sparse cone penetration tests (CPTs), we first reconstruct two-dimensional sections without geotechnical fine-tuning or predefined layer boundaries.
We evaluate the approach on synthetic sections and at two field sites in the Netherlands, comparing it with SchemaGAN, a conditional generative adversarial network trained on synthetic geotechnical sections, and with reference interpolation methods.
In the baseline comparisons, TabPFN achieved mean errors close to SchemaGAN's on synthetic sections and lower mean errors at both field sites.
On synthetic sections, TabPFN's mean absolute error decreased at each tested increase in CPT count, and, under the baseline input setting, its pointwise prediction intervals had mean coverage close to nominal levels.
Subsequent fine-tuning on the synthetic sections used to train SchemaGAN improved synthetic test accuracy but increased mean errors at both field sites.
These findings indicate that general-purpose tabular foundation models offer a promising route for updating subsurface predictions from sparse and evolving site-investigation data without retraining.
\end{abstract}


\end{frontmatter}

\section{Introduction}
\label{sec:intro}

Geotechnical engineers need to characterise subsurface stratigraphy and spatially varying ground properties to inform design and construction~\cite{baecher2003reliability}.
Phoon and Kulhawy~\cite{phoon1999characterization} described inherent soil variability in terms of its magnitude and spatial scale and distinguished it from measurement error and transformation uncertainty.
Otake and Honjo~\cite{otake2022challenges} emphasised the need to examine how uncertainty in ground conditions propagates to predictions of structural performance.
To support such assessments, engineers require spatial predictions of ground conditions between investigation locations, together with the uncertainty of these predictions.
Site-investigation data are often multivariate, sparse and incomplete, making it difficult to estimate spatial correlation structures or assess assumptions such as stationarity~\cite{phoon2022challenges}.
Cone penetration tests (CPTs), for example, provide detailed vertical profiles but only at a limited number of locations, leaving layer geometry and within-layer variability between profiles to be inferred.
Engineers therefore draw on assumptions about spatial dependence, geological knowledge and data from other sites to interpret ground conditions between observations.

Researchers have addressed the sparsity of site observations by modelling spatial dependence and incorporating data from other sites.
Geostatistical methods such as Kriging represent spatial dependence through a specified covariance model, with parameters typically estimated from the available observations.
Yoshida et al.~\cite{yoshida2021trend} used Gaussian process regression to estimate trend and random components of soil spatial variability.
Zhao and Wang~\cite{zhao2020interpolation} combined Bayesian interpolation of a multilayer soil-property profile with clustering of the interpolated profile to identify stratification from sparse measurements.
For multivariate soil-property estimation, Ching et al.~\cite{ching2021hierarchical} developed a hierarchical Bayesian model that combines site observations with information from a database of multiple sites while accounting for differences among sites.
Using this framework, Saito et al.~\cite{saito2025site} compared local and global clay databases for soil-property estimation and argued that the definition of a site in such databases should account for the complexity of vertical stratigraphy.
Engineers applying spatial models need to assess the suitability of the assumed spatial correlation structure, while those using data from other sites need to evaluate the relevance of that information to the target site~\cite{phoon2022challenges}.

Hu and Wang~\cite{hu2020probabilistic} combined random-field modelling with Monte Carlo simulation to infer soil classes and stratification in two-dimensional sections from limited CPT soundings, accounting for both interpolation and classification-model uncertainty.
Shi and Wang~\cite{shi2021development} used iterative convolution XGBoost to infer geological cross-sections from limited boreholes and a training section representing similar geological conditions.
Other approaches use deep-learning models to learn spatial patterns from collections of training sections, so the amount and representativeness of the training data become central considerations at a new site.
Campos Montero et al.~\cite{camposmontero2025schemagan} introduced SchemaGAN, a conditional generative adversarial network trained on synthetic geotechnical sections for subsurface schematisation, and evaluated it on synthetic sections and at two field sites.
The authors noted that its performance depends on the layering patterns and complexity represented in the synthetic training database.

Foundation models rely on pretraining, using broad data distributions to support a range of downstream tasks~\cite{bommasani2021foundation}.
The Tabular Prior-data Fitted Network (TabPFN) is a Transformer-based foundation model for tabular prediction that, like SchemaGAN, is pretrained on synthetic data, but on general tabular prediction tasks rather than geotechnical sections~\cite{hollmann2025tabpfn,grinsztajn2026tabpfn3}.
We use TabPFN-3, which is pretrained on synthetic tabular datasets generated from a prior that includes spatial relationships~\cite{grinsztajn2026tabpfn3}.
At inference, the model returns predictive distributions for query targets conditioned on the observed feature--target pairs supplied as context, without updating its pretrained weights~\cite{grinsztajn2026tabpfn3}.
This use of in-context learning, referred to here as in-context inference, allows the same pretrained regressor to be used with different features and targets.
Civil-engineering applications of TabPFN include estimation of pavement roughness~\cite{pang2026roughness} and prediction of pavement condition states~\cite{mizutani2026pavement}.
Geotechnical applications include estimation of rock properties and horizontal stresses~\cite{xiang2026geomechanical}, as well as spatial prediction of undrained shear strength and imputation of missing mechanical parameters~\cite{saito2026sitecharacterization}.
Related work has explored the interpretability of TabPFN predictions through embeddings and feature attribution, with a view to supporting future geotechnical decision-making~\cite{saito2026interpretable}.
In digital soil mapping, Schmidinger et al.~\cite{schmidinger2026kriging} evaluated TabPFN by augmenting environmental covariates with spatial features derived from ordinary Kriging, assessing both point and probabilistic predictions.
In an extended benchmark, they also evaluated geographic coordinates as additional features alongside those covariates.

Building on previous TabPFN applications to property prediction and imputation, we examine whether general tabular pretraining can also support subsurface schematisation.
Following the terminology of Campos Montero et al.~\cite{camposmontero2025schemagan}, we use this term for the reconstruction of two-dimensional subsurface sections from sparse CPT profiles.
This study focuses on reconstructing a continuous field of the soil behaviour type index $I_\mathrm{c}$; delineating soil units and assigning engineering properties are subsequent steps beyond its scope.
For this reconstruction, we deliberately adopt a minimal input representation based on grid coordinates and observed $I_\mathrm{c}$ values alone, although geological interpretations, borehole logs and geophysical measurements could be supplied as additional features or context observations.
For TabPFN, horizontal and vertical grid coordinates are the only features, observed $I_\mathrm{c}$ values are the context targets, and the coordinates of all grid locations in each section are the queries.
No geological labels or predefined layer boundaries are supplied, and no covariance model is explicitly prescribed.
If effective for this task, TabPFN could offer an additional option for subsurface schematisation alongside conventional interpolation and task-specific deep learning.
By adding new CPT observations to the context, we can update predictions and their uncertainty without updating the pretrained weights, potentially supporting sequential site investigation.

Our contribution is a comparative evaluation of two-dimensional $I_\mathrm{c}$ reconstruction from sparse CPT profiles, using grid coordinates as TabPFN's only features and the released SchemaGAN model as the principal comparator.
We include Kriging and nearest-neighbour interpolation as reference methods.
From the SchemaGAN study, we use the synthetic sections, generated using GeoSyn~\cite{camposmontero2025geosyn}, and the field CPT data~\cite{camposmontero2025schemagan,camposmontero2024data}, with input observations and evaluation pixels matched across methods in the accuracy and prediction-interval comparisons.
The comparison evaluates complete prediction procedures with different priors, including SchemaGAN's task-specific geotechnical prior and TabPFN's general tabular prior.
Because the two models differ in their input representations and inference procedures, their relative errors do not isolate the effect of either the prior or the network architecture.
With both neural models' released pretrained weights fixed, we first assess accuracy on synthetic sections and at held-out CPT locations in the field sections.
On synthetic sections, we extend this assessment by varying the number of input CPTs and evaluating pointwise prediction intervals, an aspect not assessed in the original SchemaGAN study.
At each field site, we examine how held-out errors vary across random input placements in the two sections used for comparison.
To illustrate flexibility in spatial extent and context size, we reconstruct the two sections at each site jointly with TabPFN using their combined input CPTs.
In a separate experiment, we test whether updating TabPFN's weights through fine-tuning on the synthetic sections used to train SchemaGAN improves prediction on both datasets.
As an initial proof of concept, this study examines the potential of general tabular pretraining for subsurface schematisation using one synthetic dataset and two field sites.
Finally, we discuss these results in terms of transfer from general tabular pretraining, sequential site investigation, and future ways to incorporate geotechnical information through context or fine-tuning.

\section{Methods}
\label{sec:method}

\subsection{Problem formulation}
\label{sec:task}

\begin{figure*}[tbp]
  \centering
  \paperfigure{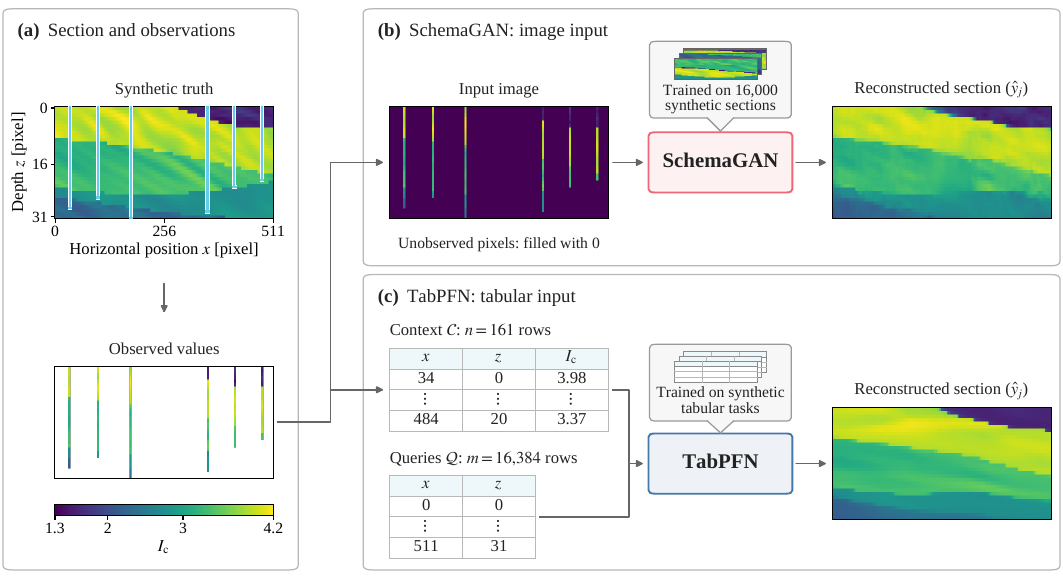}{\textwidth}{98mm}
  \caption{Input representations of SchemaGAN and TabPFN for subsurface schematisation from six input columns.
    (a) Synthetic test section and sampled observations; cyan bars mark the observed depth ranges.
    (b) Image input and reconstruction with SchemaGAN; unobserved pixels are filled with zero to mark missing input, and the input image is then normalised as described in Section~\ref{sec:baselines}.
    (c) Tabular input and reconstruction with TabPFN; the context set $\mathcal C$ contains the observed $(x,z,I_\mathrm{c})$ rows, and the query set $\mathcal Q$ contains only the $(x,z)$ coordinates of all grid pixels, including input locations.
    Both reconstructed sections show point predictions $\hat y_j$ at the grid locations.
    The vertical scale and input-column widths are exaggerated for clarity.}
  \label{fig:problem_setup}
\end{figure*}

We formulate subsurface schematisation as the reconstruction of the soil behaviour type index $I_\mathrm{c}$ over a two-dimensional vertical section from sparse CPT observations.
Each section is represented on a grid of $m$ pixels.
Let $\mathbf{s}_j=(x_j,z_j)$ denote the horizontal and vertical grid coordinates of pixel $j$, and let $y_j=I_{\mathrm{c},j}$ denote its target value, for $j=1,\ldots,m$.
The CPTs supplied as input occupy a limited number of grid columns (the input columns) and may cover different depth ranges; we refer to one such configuration as an input placement.
For a given input placement, let $\mathcal O\subseteq\{1,\ldots,m\}$ denote the index set of the valid input pixels, that is, the pixels in the input columns that hold an observed $I_\mathrm{c}$ value.
We define the context set $\mathcal C$ and the query set $\mathcal Q$ as
\begin{equation}
  \begin{aligned}
    \mathcal C &= \{(\mathbf{s}_i,y_i):i\in\mathcal O\},\\
    \mathcal Q &= \{\mathbf{s}_j:j=1,\ldots,m\}.
  \end{aligned}
  \label{eq:context_query}
\end{equation}
The context contains $n=|\mathcal O|=|\mathcal C|$ rows, each holding the coordinates and observed $I_\mathrm{c}$ value of one valid input pixel.
The query set $\mathcal Q$ covers every grid location, including those represented in $\mathcal C$.
Given $\mathcal C$, each method returns a point prediction $\hat y_j$ of $y_j$ at every $\mathbf{s}_j\in\mathcal Q$.
The pixels used to evaluate the predictions are specified for each experiment in Section~\ref{sec:protocol}.

Figure~\ref{fig:problem_setup} illustrates the subsurface schematisation task and the input representations used by SchemaGAN and TabPFN.
SchemaGAN receives the values in $\mathcal C$ as an image, with each value placed at its grid location and the unobserved pixels, whose indices are outside $\mathcal O$, set to zero (Figure~\ref{fig:problem_setup}b).
TabPFN receives the same $\mathcal C$ as tabular rows and predicts $I_\mathrm{c}$ at the locations in $\mathcal Q$ (Figure~\ref{fig:problem_setup}c).
The datasets and preprocessing are described in Section~\ref{sec:data}, the prediction methods in Sections~\ref{sec:tabpfn} and~\ref{sec:baselines}, and the experimental design and evaluation in Section~\ref{sec:protocol}.

\subsection{Datasets and preprocessing}
\label{sec:data}

\begin{figure*}[tbp]
  \centering
  \paperfigure{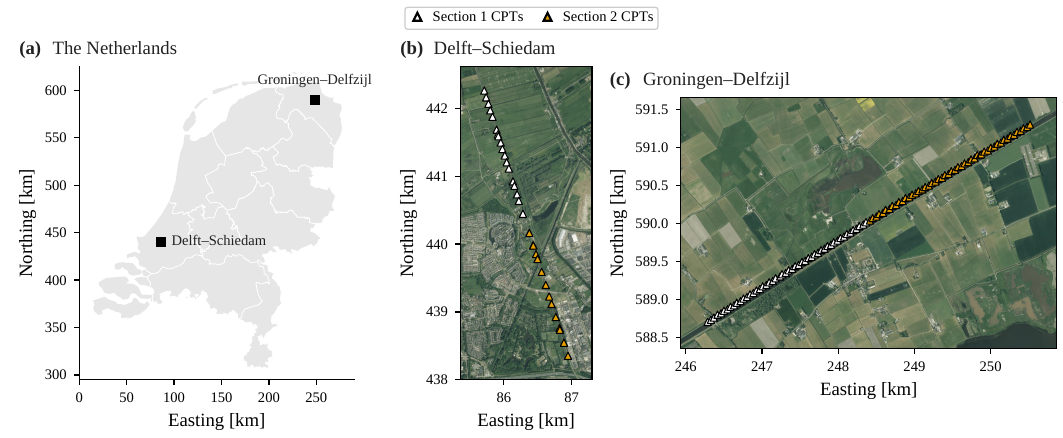}{\textwidth}{70mm}
  \caption{Locations of the field CPTs and their assignment to the two sections at each site.
    Panel (a) shows the locations of the two sites in the Netherlands, and panels (b) and (c) show CPT locations overlaid on aerial photographs.
    In panels (b) and (c), white and orange triangles mark the CPTs in sections 1 and 2, respectively.
    Axes show RD New (EPSG:28992) coordinates in kilometres.
    Aerial photographs (2025): Beeldmateriaal Nederland; province boundaries: Kadaster; both obtained via PDOK under CC BY 4.0 (\url{https://creativecommons.org/licenses/by/4.0/}).
    Land and lakes: Natural Earth (public domain).}
  \label{fig:field_section_layout}
\end{figure*}

We evaluate TabPFN and the comparison methods on the subsurface reconstruction task studied by Campos Montero et al.~\cite{camposmontero2025schemagan}, using the released synthetic dataset and the preprocessed field sections from two sites in the Netherlands.
Each synthetic section represents up to five layers on a grid of 32 depth rows and 512 horizontal columns (Figure~\ref{fig:problem_setup}a).
Layer boundaries follow randomly parameterised periodic functions, and each layer is assigned one of five soil types.
Within each layer, $I_\mathrm{c}$ values are generated using an anisotropic Gaussian random field with mean and standard deviation determined by the assigned soil type~\cite{camposmontero2025geosyn}.
The released dataset~\cite{camposmontero2024data} comprises 16{,}000 training, 4{,}000 validation and 4{,}000 test sections generated using the same procedure.
SchemaGAN was trained on the training split; in the main experiments, we use TabPFN without geotechnical fine-tuning.
We use the released test split for the synthetic comparisons and the training and validation splits for TabPFN fine-tuning and checkpoint selection, respectively (Section~\ref{sec:experiments}).

We use the preprocessed field sections from the original study unchanged, with two $32\times512$ grids per site (Figure~\ref{fig:field_section_layout}).
The original preprocessing included a top cutoff to create a flat surface and averaging of $I_\mathrm{c}$ within the depth intervals represented by the 32 rows.
Sections 1 and 2 at Delft--Schiedam contain 18 and 14 CPTs, respectively (32 in total), and those at Groningen--Delfzijl contain 50 and 51 CPTs, respectively (101 in total).
Each CPT occupies a separate grid column, termed a measured column, with contiguous measurements from the top row to its termination depth; in Delft--Schiedam section 1, no CPT reaches the bottom.
Zero entries mark missing measurements and are excluded from the input observations.

\subsection{Spatial prediction with TabPFN}
\label{sec:tabpfn}

TabPFN is a Transformer-based foundation model for tabular prediction~\cite{hollmann2025tabpfn}.
We use TabPFN-3, pretrained exclusively on synthetic tabular datasets generated from a structural causal model prior, with a reported training scale exceeding eight trillion tokens.
The prior covers diverse relationships among variables, including spatial relationships, but the TabPFN-3 report describes no pretraining task specific to geotechnical sections~\cite{grinsztajn2026tabpfn3}.
During pretraining, the model learns to predict targets for query rows from context rows with known features and targets.
At inference, TabPFN predicts targets from the supplied context without updating its weights.

For each grid and input placement, we supply the grid coordinates $(x,z)$ as the only features and $I_\mathrm{c}$ as the context target (Figure~\ref{fig:problem_setup}c).
Each valid input pixel contributes one context row, giving $n\leq32K$ rows for $K$ input columns (at most 192 for six input columns).
We use zero-based column and row indices as $x$ and $z$, with $x$ increasing to the right and $z$ increasing downwards from the top-left pixel.
For a $32\times512$ section, $x\in\{0,\ldots,511\}$ and $z\in\{0,\ldots,31\}$, giving $m=16{,}384$ query locations.
For reconstruction, we query all grid locations in $\mathcal Q$ and score predictions on the pixels specified in Section~\ref{sec:protocol}.
In all experiments, including fine-tuning, we use these grid indices without supplying physical grid spacings or section aspect ratios, and we do not assign physical dimensions to the synthetic sections.

We use the released TabPFN-3 default regression weights, identified by the checkpoint {\def\UrlBreaks{\do\-\do\_\do\.}\nolinkurl{tabpfn-v3-regressor-v3_default.ckpt}}, except when evaluating the fine-tuned models.
For reconstruction, we use the TabPFN package's inference-time ensemble of eight members that share the same network weights, with inference seed 0 (\texttt{tabpfn} version 8.5.0).
We retain TabPFN's default preprocessing, including feature-wise standardisation.
Each member applies one of the four combinations of feature and target preprocessing supplied with the released model and a random permutation of the preprocessed features.
The predictive distribution of each member is computed with the package's default softmax temperature of 0.9, and the eight distributions are averaged to obtain $p_\theta(y_j\mid\mathbf{s}_j,\mathcal C)$ in the original $I_\mathrm{c}$ units, where $\theta$ denotes the model weights.

Let $q_\tau(\mathbf{s}_j;\mathcal C)$ denote the $\tau$ quantile of this distribution.
The point prediction is its median,
\begin{equation}
  \hat y_j=q_{0.5}(\mathbf{s}_j;\mathcal C),
  \label{eq:tabpfn_point}
\end{equation}
which we use for both the mean absolute error (MAE) and the root mean squared error (RMSE).
For the comparison of predictive medians and means in Section~\ref{sec:baselines}, we obtain both summaries from the averaged predictive distribution and use the predictive mean for RMSE.
Let $\alpha\in(0,1)$ denote the nominal probability that the target value lies outside the prediction interval, so that $1-\alpha$ is its nominal coverage.
A nominal $100(1-\alpha)\%$ pointwise prediction interval (PI) is
\begin{equation}
  \mathrm{PI}_{1-\alpha}(\mathbf{s}_j)
  =\left[q_{\alpha/2}(\mathbf{s}_j;\mathcal C),\ q_{1-\alpha/2}(\mathbf{s}_j;\mathcal C)\right].
  \label{eq:tabpfn_interval}
\end{equation}
For example, $\alpha=0.10$ gives a nominal 90\% prediction interval bounded by the 0.05 and 0.95 quantiles.

\subsection{Comparison methods}
\label{sec:baselines}

We focus on comparing TabPFN and SchemaGAN, which have different pretraining tasks.
SchemaGAN incorporates a task-specific prior learned from synthetic geotechnical sections~\cite{camposmontero2025schemagan}, whereas TabPFN incorporates a general tabular prior~\cite{grinsztajn2026tabpfn3}.
Both models receive the observations in $\mathcal C$, with spatial positions encoded by pixel locations in the input image for SchemaGAN and supplied as grid-coordinate features for TabPFN.
Their pretraining data, input representations and inference procedures differ, so the comparison evaluates the complete prediction procedures and does not isolate the effect of either the prior or the network architecture.
We include Kriging and Nearest Neighbour, the two interpolation methods compared with SchemaGAN at the field sites in the original study~\cite{camposmontero2025schemagan}.

SchemaGAN is a conditional generative adversarial network based on pix2pix~\cite{isola2017pix2pix} that reconstructs a complete $I_\mathrm{c}$ section from a sparse CPT-like input image~\cite{camposmontero2025schemagan}.
Its modified U-Net generator was trained against a patch-based discriminator using adversarial and $L_1$ reconstruction losses.
The released training implementation uses six input columns per section~\cite{schemagan_source}.
We retain the released pretrained weights~\cite{camposmontero2024data} and $32\times512$ image dimensions throughout the experiments.
Following the released implementation~\cite{schemagan_source}, we apply the same linear normalisation used during pretraining to the input image after the unobserved pixels have been set to zero, and map model outputs back to the original $I_\mathrm{c}$ units.
In the released generator, dropout remains active during inference and batch normalisation uses the statistics of the current batch.
As in the released evaluation code, each input image is processed in a separate forward pass with batch size 1.

Unless stated otherwise, we use 100 SchemaGAN reconstructions per input placement, generated using Monte Carlo dropout~\cite{gal2016dropout} with inference seeds 0--99.
We score each reconstruction separately and average the resulting errors to reduce dependence on a single dropout realisation (Section~\ref{sec:metrics}).
To examine how the choice of point prediction affects the comparison, we also evaluate the predictive medians of both models using MAE and their predictive means using RMSE.
For SchemaGAN, we use the pixelwise sample median and mean of these reconstructions.
We assess these summaries for pointwise accuracy; they need not preserve the spatial structure of an individual generated reconstruction.
We also use the same generated reconstructions to construct pointwise prediction intervals, which were not evaluated in the original study~\cite{camposmontero2025schemagan}.
The released SchemaGAN model was trained for reconstruction, and we apply no calibration procedure targeting nominal coverage to the intervals constructed here.
At nominal level $100(1-\alpha)\%$, we use a normal approximation at each pixel to define the interval as the sample mean plus or minus $\Phi^{-1}(1-\alpha/2)$ times the sample standard deviation across these reconstructions, where $\Phi^{-1}$ is the standard normal quantile function.

Kriging and Nearest Neighbour follow the implementations released with the original study~\cite{camposmontero2025schemagan,schemagan_source}, using the same grid coordinates as TabPFN.
The Kriging implementation uses Gaussian process regression with a covariance given by the sum of an anisotropic radial basis function kernel and a white-noise kernel.
For each fit, we centre and scale the input $I_\mathrm{c}$ values using their mean and standard deviation and transform predictions back to $I_\mathrm{c}$ units; the Gaussian process has no spatial trend term.
The kernel parameters are fitted separately for each input placement by maximising the log marginal likelihood with the L-BFGS-B optimiser and ten random restarts.
The two length scales and the noise variance are bounded between $10^{-5}$ and $10^{5}$, and the seed of the random restarts is fixed to the zero-based index assigned to each combination of section and input placement.
We use the posterior mean for point prediction.
We construct central pointwise prediction intervals from the Gaussian posterior predictive distribution, including the white-noise contribution in the predictive variance.
The median fitted horizontal and vertical length scales were 136 and 2.5 pixels on the synthetic sections, 67 and 1.6 pixels at Delft--Schiedam and 54 and 1.5 pixels at Groningen--Delfzijl.
Of the 8{,}000 fits, 72 produced warnings, including one L-BFGS-B convergence warning; all other warnings concerned parameters close to their bounds, and all predictions were retained.
Comparisons with Kriging refer to this particular covariance and fitting procedure.
Nearest Neighbour assigns each pixel the value of the nearest observed pixel, using Euclidean distance in grid coordinates, and provides point predictions only.

\subsection{Experimental design and evaluation}
\label{sec:protocol}

For each input placement in the accuracy and prediction-interval comparisons, the methods being compared received the same input observations $\mathcal C$ and were evaluated on the same pixels.
After inference, predictions were neither clipped to a prescribed $I_\mathrm{c}$ range nor replaced with observed values at input pixels.

\subsubsection{Experimental setup}
\label{sec:experiments}

The baseline accuracy comparisons on both datasets included all four methods.
For each synthetic section, the baseline input was drawn by the released sampling procedure~\cite{schemagan_source} and comprised $K=6$ randomly selected input columns subject to its minimum-spacing constraint, with randomly truncated depths.
These inputs covered approximately 1\% of the grid pixels on average.
We evaluated prediction errors over all $m$ grid pixels in $\mathcal Q$, including the input pixels~\cite{schemagan_source} (Section~\ref{sec:res_synthetic}).
With the same inputs, we analysed TabPFN and SchemaGAN errors at unobserved pixels by Euclidean distance in grid coordinates to the nearest observed pixel (Section~\ref{sec:res_conditions}).
Using the baseline six-CPT inputs, we evaluated prediction intervals from TabPFN, SchemaGAN and Kriging over unobserved pixels at nominal levels of 50, 80, 90 and 95\%, without additional calibration (Section~\ref{sec:res_intervals}).

\begingroup
\emergencystretch=6em
To assess how the released models respond to changes in CPT count without retraining, we evaluated TabPFN and SchemaGAN with \mbox{$K\in\{3,6,12,24,48\}$} input columns (Section~\ref{sec:res_conditions}).
For SchemaGAN, counts other than six differ from its released training configuration.
We used each synthetic section's baseline input for $K=6$ and a random subset of three of its columns for $K=3$.
Starting from the six-column input, we successively added randomly selected unused columns to obtain the 12-, 24- and 48-column inputs, without a minimum-spacing constraint.
Existing observations were retained, and added columns used the same random depth truncation as the baseline inputs.
In each section, predictions were scored on the pixels unobserved in all five input placements.
\par\endgroup
For the baseline field accuracy comparisons (Section~\ref{sec:res_field}), we drew 1{,}000 input placements per section, each comprising $K=6$ measured columns.
Following the original study's procedure for selecting field inputs~\cite{camposmontero2025schemagan,schemagan_source}, we drew columns sequentially at random subject to its minimum-spacing constraint, retaining repeated six-column combinations as separate input placements.
All remaining measured columns were held out for evaluation regardless of their distance from the input columns.
For both datasets, the comparison of predictive medians and means (Section~\ref{sec:baselines}) used the same inputs and evaluation pixels as the baseline comparisons.

For the field reconstruction examples in Section~\ref{sec:res_fullsite}, we reconstructed each site's two $32\times512$ sections independently with SchemaGAN and TabPFN, using six input CPTs per section.
For joint reconstruction, we placed the two original section grids side by side without rescaling to form a $32\times1{,}024$ grid, assuming that equal row indices represented the same depth interval.
TabPFN reconstructed this joint grid using the combined observations from the 12 input CPTs.
The split and joint TabPFN reconstructions used the same model weights and inference settings.
No boundary smoothing was applied to any split or joint reconstruction.

The preceding experiments used TabPFN without training it specifically on geotechnical sections.
To assess the effect of such training (Section~\ref{sec:res_finetune}), we fine-tuned all TabPFN weights $\theta$ on the same 16{,}000 synthetic training sections used to train SchemaGAN.
At each training step, we used $\mathcal C$ as context and sampled prediction targets at unobserved locations.
The loss was the sum of the continuous ranked probability score (CRPS) and the mean squared error (MSE) of the predictive mean, both computed in TabPFN's internal target space.
We trained for five epochs with optimisation seed 0.
We selected the checkpoint with the lowest mean MAE over unobserved pixels, using only the original study's 4{,}000 synthetic validation sections~\cite{camposmontero2025schemagan}.
The selected checkpoint was evaluated on the synthetic test and field sections using the inference settings in Section~\ref{sec:tabpfn}.
Training details are given in \ref{app:finetuning}.
To characterise how the synthetic training sections differ from the field sections, we also compared three descriptors of $I_\mathrm{c}$: its mean, the mean absolute difference between adjacent depth rows, and the mean absolute difference between neighbouring measured columns at the same depth row.
For each field section, we calculated these descriptors at its CPT locations and depth rows and at the same grid positions in each of the 16{,}000 synthetic training sections.

\subsubsection{Metrics and statistical comparison}
\label{sec:metrics}

We assessed each method's point predictions $\hat y_j$ against target values $y_j$ using MAE and RMSE in $I_\mathrm{c}$.
Following the original evaluation code~\cite{camposmontero2025schemagan,schemagan_source}, field errors in each held-out column were averaged over all 32 rows, with zero residuals assigned below the CPT's termination depth.
Synthetic errors were calculated per section; field errors were calculated per held-out column and averaged equally across columns within each input placement.
For field RMSE, the square root was taken within each held-out column before averaging across columns.
Unless stated otherwise, SchemaGAN errors were averaged across the 100 reconstructions for each input placement before calculating the statistical summaries.

The accuracy tables in Sections~\ref{sec:res_synthetic} and~\ref{sec:res_field} report means and sample standard deviations across the 4{,}000 synthetic sections or the 2{,}000 input placements per field site.
We also report the percentage of synthetic sections or field input placements in which each method had lower error than SchemaGAN.
For these tables, we paired errors by synthetic section or by input placement within each field section and calculated mean differences (method minus SchemaGAN) with normal-approximation 95\% confidence intervals.
Field standard errors were based on variation in paired differences within each section, with equal weight given to the two section means.
Field confidence intervals quantify input-placement uncertainty conditional on the two observed sections and inference seeds.
For the CPT-count and distance curves, pointwise 95\% confidence intervals used the same normal approximation, with standard errors estimated across contributing sections.
In all box plots, boxes span the first to third quartiles and the central lines indicate medians.
To complement MAE and RMSE, we calculated ratios of the mean and median absolute $I_\mathrm{c}$ differences in each reconstructed section to the corresponding values in its target section, using all pixel pairs separated vertically by 1, 2, 4 or 8 rows, or horizontally by 1, 2, 4, 8, 16, 32, 64 or 128 columns.
Ratios below one indicate smaller mean or median absolute differences than in the target section at the same separation.
Using the baseline six-CPT reconstructions, we calculated ratios separately for each section and separation and report medians across the 4{,}000 synthetic test sections.
For SchemaGAN, we first averaged the ratios from 100 individually evaluated reconstructions within each section.

In addition to evaluating point predictions, we assessed pointwise prediction intervals $\mathrm{PI}_{1-\alpha}(\mathbf{s}_j)$ using empirical coverage, mean width and interval score.
For each synthetic section $s$, let $\mathcal U_s$ be the set of unobserved pixel indices.
For each method at a fixed nominal level $1-\alpha$, write $[L_{sj},U_{sj}]$ for the interval at pixel $j$, constructed as described in Sections~\ref{sec:tabpfn} and~\ref{sec:baselines}, and $y_{sj}$ for its target value.
The empirical coverage and mean interval width in section $s$ are
{\color{black}
\begin{align}
  \widehat C_s &= \frac{1}{|\mathcal U_s|}\sum_{j\in\mathcal U_s}\mathbf{1}\{L_{sj}\leq y_{sj}\leq U_{sj}\},
  \label{eq:interval_coverage}\\
  \overline W_s &= \frac{1}{|\mathcal U_s|}\sum_{j\in\mathcal U_s}(U_{sj}-L_{sj}),
  \label{eq:interval_mean_width}
\end{align}
}
where $\mathbf{1}\{\cdot\}$ equals one if its condition is satisfied and zero otherwise.
The interval score combines interval width with a penalty for targets outside the interval~\cite{gneiting2007scoring}.
For each pixel, it is
{\color{black}
\begin{equation}
  \mathrm{IS}_{\alpha,sj}=(U_{sj}-L_{sj})+\frac{2}{\alpha}(L_{sj}-y_{sj})\mathbf{1}\{y_{sj}<L_{sj}\}+\frac{2}{\alpha}(y_{sj}-U_{sj})\mathbf{1}\{y_{sj}>U_{sj}\}.
  \label{eq:interval_score}
\end{equation}
}
We obtain the section mean $\overline{\mathrm{IS}}_{\alpha,s}=|\mathcal U_s|^{-1}\sum_{j\in\mathcal U_s}\mathrm{IS}_{\alpha,sj}$.
We average $\widehat C_s$, $\overline W_s$ and $\overline{\mathrm{IS}}_{\alpha,s}$ equally across all 4{,}000 synthetic test sections, with coverage reported as a percentage.
For the example sections in Section~\ref{sec:res_intervals}, we report $\widehat C_s$ and $\overline W_s$ at $\alpha=0.10$.

\section{Results}
\label{sec:results}

In the main experiments, we observed three principal findings.
First, without geotechnical fine-tuning, TabPFN achieved mean errors close to those of SchemaGAN on synthetic test sections and the lowest mean MAE and RMSE among the four methods at held-out CPT columns in both field sites in the baseline comparison (Sections~\ref{sec:res_synthetic} and~\ref{sec:res_field}).
Second, with fixed pretrained weights, TabPFN's mean MAE decreased at each tested increase in CPT count on synthetic sections, and it reconstructed the two field-section grids at each site jointly using their combined input CPTs (Sections~\ref{sec:res_conditions} and~\ref{sec:res_fullsite}).
Third, on the synthetic test sections with six input CPTs, TabPFN's pointwise prediction intervals had mean coverage close to the nominal levels and narrower mean widths than those of Kriging (Section~\ref{sec:res_intervals}).
In the fine-tuning experiment, mean MAE and RMSE decreased on synthetic test sections but increased at both field sites (Section~\ref{sec:res_finetune}).

\subsection{Prediction accuracy on synthetic sections}
\label{sec:res_synthetic}

\begin{figure}[tbp]
  \centering
  \paperfigure{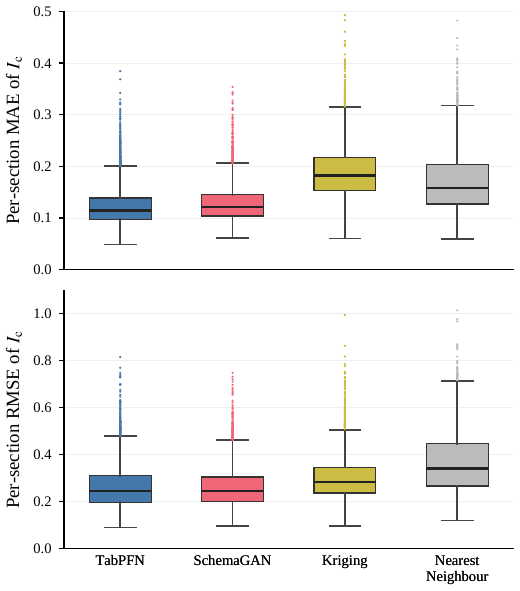}{88mm}{100mm}
  \caption{Distributions of section-wise MAE (top) and RMSE (bottom) in $I_\mathrm{c}$, computed over all pixels of each of the 4{,}000 synthetic test sections.
    Whiskers extend to the most extreme values within 1.5 interquartile ranges of the box.
    In total, 18 MAE values lie above the displayed range.}
  \label{fig:synthetic_mae_rmse_box}
\end{figure}

\begin{table}[tbp]
  \centering
  \caption{Section-wise errors in $I_\mathrm{c}$ evaluated over all pixels, including the input pixels, of the 4{,}000 synthetic test sections, reported as mean (SD) across sections.
    Comparisons with SchemaGAN are paired by section; negative differences indicate lower error.
    The final column reports the percentage of sections with lower error than SchemaGAN.
    SchemaGAN results are averaged over 100 separately scored inference seeds; TabPFN uses one seed, Kriging uses a fixed optimisation seed for each fit, and Nearest Neighbour is deterministic.}
  \label{tab:synthetic_summary}
\begin{tabular}{llrrr}
\toprule
Metric & Method & Mean (SD) & Mean difference (95\,\% CI) & \shortstack{Sections with\\lower error (\%)} \\
\midrule
MAE & TabPFN & 0.123 (0.038) & $-0.0052$ ($-0.0059$, $-0.0045$) & 63.4 \\
 & SchemaGAN & 0.128 (0.035) & --- & --- \\
 & Kriging & 0.191 (0.059) & $+0.0628$ ($+0.0614$, $+0.0641$) & 1.8 \\
 & Nearest Neighbour & 0.170 (0.058) & $+0.0417$ ($+0.0404$, $+0.0430$) & 9.5 \\
\midrule
RMSE & TabPFN & 0.264 (0.094) & $+0.0043$ ($+0.0027$, $+0.0058$) & 47.9 \\
 & SchemaGAN & 0.260 (0.085) & --- & --- \\
 & Kriging & 0.299 (0.091) & $+0.0398$ ($+0.0380$, $+0.0417$) & 17.2 \\
 & Nearest Neighbour & 0.365 (0.130) & $+0.1053$ ($+0.1024$, $+0.1081$) & 6.0 \\
\bottomrule
\end{tabular}

\end{table}

Figure~\ref{fig:synthetic_mae_rmse_box} shows the distributions of MAE and RMSE across the 4{,}000 synthetic test sections under the baseline input setting (Section~\ref{sec:experiments}), and Table~\ref{tab:synthetic_summary} reports their means and paired comparisons with SchemaGAN.
Without geotechnical fine-tuning, TabPFN achieved mean MAE and RMSE close to those of SchemaGAN, which was trained specifically on synthetic geotechnical sections, although its errors varied more across sections than SchemaGAN's, with more upper-tail outliers.
Both models achieved lower mean MAE and RMSE than Kriging and Nearest Neighbour.
Using predictive medians for both models preserved the MAE ordering shown in Table~\ref{tab:synthetic_summary}.
With predictive means, however, TabPFN had lower mean RMSE than SchemaGAN (0.2544 versus 0.2561), reversing their ordering in the baseline comparison (Section~\ref{sec:baselines}).

\begin{figure*}[tbp]
  \centering
  \paperfigure{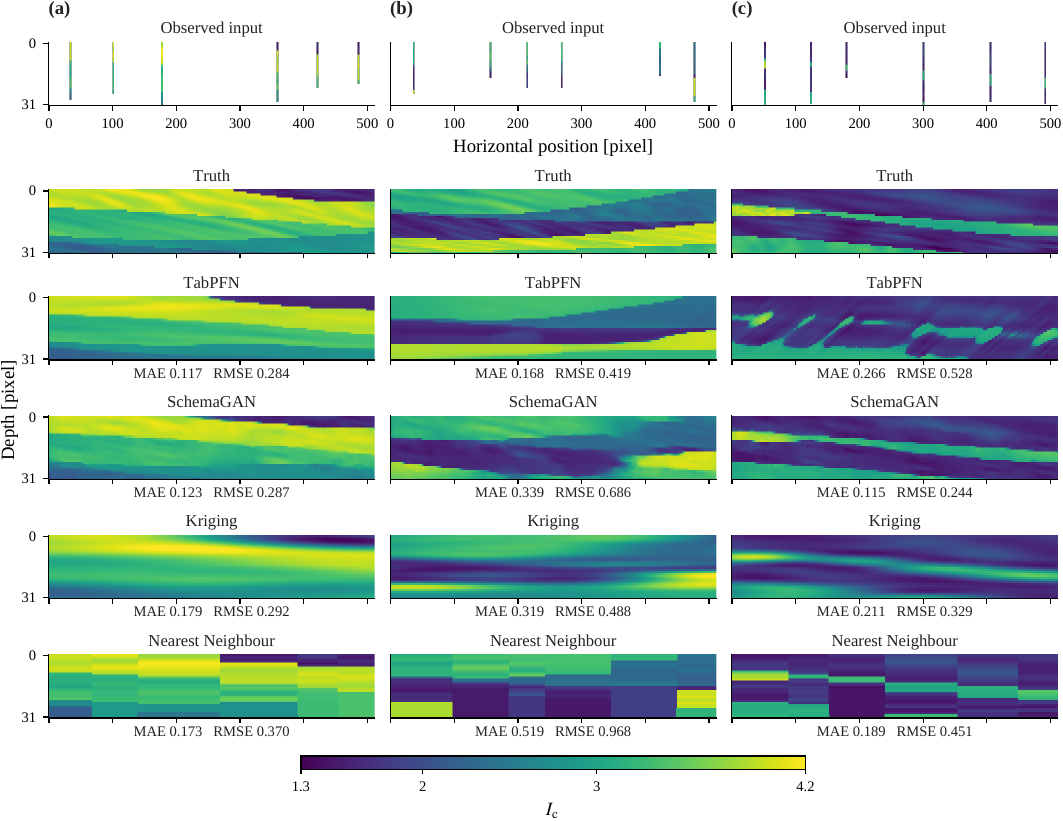}{\textwidth}{80mm}
  \caption{Reconstructions of three sections selected from the 4{,}000 synthetic test sections: (a) a representative section, (b) the section maximising SchemaGAN MAE minus TabPFN MAE, and (c) the section maximising TabPFN MAE minus SchemaGAN MAE.
    For selection, SchemaGAN MAE is averaged over 100 inference seeds; displayed SchemaGAN predictions and errors use inference seed 0.
    MAE and RMSE are computed over all pixels.
    The vertical scale and input-column widths are exaggerated for clarity.}
  \label{fig:synthetic_section_grid}
\end{figure*}

Figure~\ref{fig:synthetic_section_grid}a shows a representative section selected for a near-median within-section standard deviation of $I_\mathrm{c}$ and typical MAE across all four methods.
The other two sections maximise the MAE difference computed as SchemaGAN minus TabPFN (b) and TabPFN minus SchemaGAN (c), respectively.
In the representative section, TabPFN and SchemaGAN both recovered the main layer geometry, although TabPFN reproduced less of the within-layer variation than SchemaGAN (Figure~\ref{fig:synthetic_section_grid}a).
Kriging produced diffuse layer transitions, and Nearest Neighbour introduced abrupt lateral steps.
TabPFN preserved the continuous high-$I_\mathrm{c}$ layer near the bottom where SchemaGAN lost its continuity (Figure~\ref{fig:synthetic_section_grid}b), whereas SchemaGAN recovered the thin inclined layer that TabPFN reconstructed as disconnected patches (Figure~\ref{fig:synthetic_section_grid}c).
Despite the small differences in mean MAE and RMSE across the synthetic test sections (Figure~\ref{fig:synthetic_mae_rmse_box}; Table~\ref{tab:synthetic_summary}), TabPFN and SchemaGAN differed in their reconstruction of layer continuity and within-layer variation in the illustrated sections.

Across all 4{,}000 synthetic test sections, we quantified spatial variation using reconstructed-to-target ratios of mean absolute $I_\mathrm{c}$ differences at each separation (Section~\ref{sec:metrics}; Table~\ref{tab:spatial_difference_ratios}).
At a vertical separation of one grid row, the across-section median ratios were 0.65 for TabPFN and 0.76 for SchemaGAN.
Across the tested horizontal separations between 1 and 128 grid columns, the corresponding medians ranged from 0.37 to 0.82 and from 0.74 to 1.23, respectively.
TabPFN's median ratios were below one at every tested separation, whereas SchemaGAN's were higher than TabPFN's at every separation and exceeded one at horizontal separations of one and two grid columns.
We observed the same pattern for the ratios of median absolute differences in the same table.

\subsection{Dependence of accuracy on CPT count and distance to observations}
\label{sec:res_conditions}

\begin{figure}[tbp]
  \centering
  \paperfigure{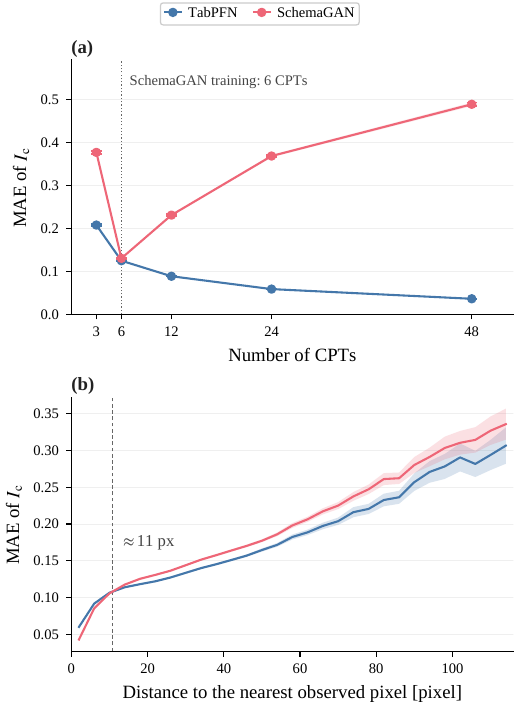}{88mm}{120mm}
  \caption{Mean MAE of $I_\mathrm{c}$ at unobserved pixels in the 4{,}000 synthetic test sections against (a) the number of CPTs and (b) the Euclidean distance in grid coordinates to the nearest observed pixel.
    In (a), model weights and unobserved evaluation pixels were fixed across CPT counts; the dotted line marks SchemaGAN's released six-CPT training configuration.
    In (b), the baseline six-CPT input is used; only 4-pixel distance bins represented in at least 200 sections (5\%) are shown.
    Error bars and shading show pointwise 95\% confidence intervals for the mean across contributing sections.
    The dashed line in (b) marks the interpolated crossing of the mean curves (approximately 11 pixels).}
  \label{fig:observation_conditions}
\end{figure}

Figure~\ref{fig:observation_conditions}a shows mean MAE across the 4{,}000 synthetic test sections for different numbers of input CPTs, with both models' pretrained weights held fixed (Section~\ref{sec:experiments}).
For TabPFN, mean MAE decreased with every increase in CPT count from 3 to 48.
SchemaGAN achieved its lowest mean MAE at six CPTs, matching the input count in its released training implementation.
The difference between the mean curves was smallest at six CPTs.
Above six CPTs, SchemaGAN's mean MAE increased as more input columns were added.
This increase does not establish an inherent inability of the SchemaGAN architecture to use additional observations.
Figure~\ref{fig:observation_conditions}b shows that mean MAE generally increased with distance to the nearest observed pixel for both methods under the baseline six-CPT input setting.
SchemaGAN had lower mean MAE near the observed pixels, whereas TabPFN had lower mean MAE at greater distances, with the mean curves crossing at approximately 11 pixels.
Local reconstruction differences and the relation to training observation conditions are discussed in Sections~\ref{sec:disc_transfer} and~\ref{sec:disc_incontext}, respectively.

\subsection{Calibration and sharpness of prediction intervals}
\label{sec:res_intervals}

\begin{table}[tbp]
  \centering
  \caption{Pointwise prediction interval performance for $I_\mathrm{c}$ over the 4{,}000 synthetic test sections.
    Metrics are computed over unobserved pixels within each section and averaged equally across sections.
    SchemaGAN intervals were constructed in this study using Monte Carlo dropout and a pixelwise normal approximation (Section~\ref{sec:baselines}).
    Lower interval scores indicate better performance at a given nominal level.}
  \label{tab:interval_summary}
\begin{tabular}{clrrr}
\toprule
Nominal level (\%) & Method & Coverage (\%) & Mean width & Interval score \\
\midrule
50 & TabPFN & 51.6 & 0.219 & 0.412 \\
 & SchemaGAN (MC dropout) & 5.7 & 0.025 & 0.489 \\
 & Kriging & 68.5 & 0.393 & 0.689 \\
\midrule
80 & TabPFN & 79.9 & 0.440 & 0.636 \\
 & SchemaGAN (MC dropout) & 10.9 & 0.048 & 1.107 \\
 & Kriging & 86.1 & 0.747 & 1.108 \\
\midrule
90 & TabPFN & 89.2 & 0.597 & 0.822 \\
 & SchemaGAN (MC dropout) & 13.9 & 0.062 & 2.071 \\
 & Kriging & 90.8 & 0.959 & 1.441 \\
\midrule
95 & TabPFN & 94.1 & 0.762 & 1.039 \\
 & SchemaGAN (MC dropout) & 16.6 & 0.074 & 3.915 \\
 & Kriging & 93.5 & 1.142 & 1.825 \\
\bottomrule
\end{tabular}

\end{table}

Table~\ref{tab:interval_summary} summarises interval performance across all 4{,}000 synthetic test sections (Section~\ref{sec:experiments}).
TabPFN's mean coverage remained close to the nominal level at all four tested levels, while its mean interval widths were smaller than those of Kriging.
Kriging's mean coverage exceeded the nominal levels at 50\% and 80\%, reaching 68.5\% and 86.1\%, respectively.
At the nominal 90\% level, mean coverage was 89.2\% for TabPFN, 90.8\% for Kriging and 13.9\% for the Monte Carlo dropout intervals constructed here for SchemaGAN.
At this level, TabPFN's mean width was approximately 38\% smaller than Kriging's.
The SchemaGAN Monte Carlo dropout intervals had the smallest mean widths at every level, with mean coverage reaching 16.6\% at the nominal 95\% level.
At the 50\% level, the SchemaGAN dropout intervals had a lower mean interval score than Kriging's (0.489 versus 0.689), despite substantial undercoverage.
TabPFN achieved the lowest mean interval score at every nominal level.

\begin{figure*}[tbp]
  \centering
  \paperfigure{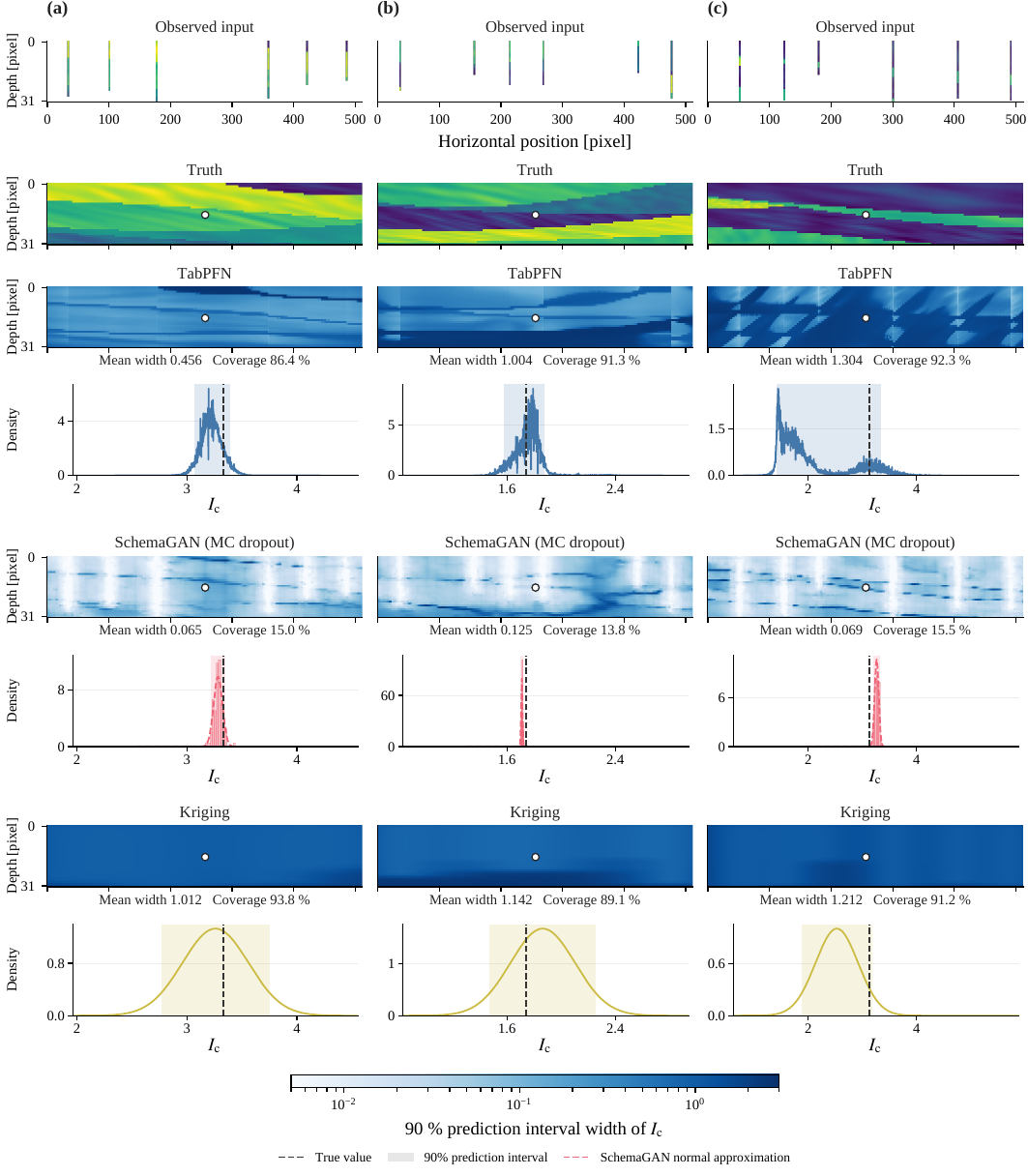}{\textwidth}{206mm}
  \caption{Pointwise 90\,\% prediction interval widths and predictive distributions for the same three synthetic sections as Figure~\ref{fig:synthetic_section_grid}.
    The observed input and truth rows use the $I_\mathrm{c}$ scale of that figure; interval-width maps share the logarithmic colour scale below.
    Density panels show the predictive distributions at the unobserved point $(x,z)=(256,16)$ marked by white circles; SchemaGAN's histogram uses 100 dropout draws.
    Mean width and coverage are computed over unobserved pixels; the maps show all pixels.
    The vertical scale and input-column widths of the maps are exaggerated for clarity.}
  \label{fig:interval_width_grid}
\end{figure*}

Figure~\ref{fig:interval_width_grid} shows the widths of the pointwise nominal 90\% prediction intervals (Section~\ref{sec:metrics}) and the predictive distributions at the marked unobserved points for the same three sections as Figure~\ref{fig:synthetic_section_grid}.
TabPFN produced spatially varying widths, with wider intervals near layer transitions and in gaps between input CPTs.
Kriging's widths varied much less than those of TabPFN, with local increases below the depths reached by shallower input CPTs in all three panels, most extensively in panel (b).
TabPFN's mean interval width was larger in panels (b) and (c) than in panel (a), and exceeded Kriging's in panel (c).
The Monte Carlo dropout intervals constructed here for SchemaGAN showed bands of increased width near layer boundaries and were much narrower than those of the other two methods.
Their coverage remained far below the nominal level in all three examples, including panel (c), where SchemaGAN had lower point-prediction MAE than TabPFN (Figure~\ref{fig:synthetic_section_grid}c).
In the density panels, TabPFN's predictive distributions at the marked points were concentrated in (a) and (b) but broad and multimodal in (c).
The intervals from TabPFN and Kriging contained the target value at all three points, whereas SchemaGAN's intervals were the narrowest and excluded it in panels (b) and (c).

\subsection{Prediction accuracy on field CPT sections}
\label{sec:res_field}

\begin{figure}[tbp]
  \centering
  \paperfigure{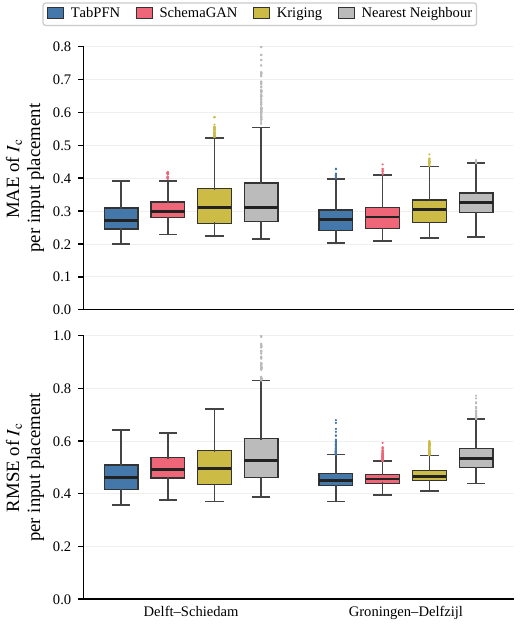}{88mm}{106mm}
  \caption{Distributions of MAE (top) and RMSE (bottom) in $I_\mathrm{c}$ per input placement, computed at held-out CPT columns for 2{,}000 input placements per site (1{,}000 on each of two sections).
    Whiskers extend to the most extreme values within 1.5 interquartile ranges of the box.
    In total, 10 RMSE values lie above the displayed range.}
  \label{fig:field_mae_rmse_box}
\end{figure}

\begin{table}[tbp]
  \centering
  \caption{Errors in $I_\mathrm{c}$ at held-out CPT columns, reported as mean (SD) over 2{,}000 input placements per site.
    Comparisons with SchemaGAN are paired by input placement; negative differences indicate lower error.
    The final column reports the percentage of input placements with lower error than SchemaGAN.
    Confidence intervals quantify input-placement uncertainty within the two fixed sections of each site.
    SchemaGAN results are averaged over 100 separately scored inference seeds for each placement; TabPFN uses one seed, Kriging uses a fixed optimisation seed for each fit, and Nearest Neighbour is deterministic.}
  \label{tab:field_summary}
\begin{tabular}{llrrr}
\toprule
Metric & Method & Mean (SD) & Mean difference (95\,\% CI) & \shortstack{Placements with\\lower error (\%)} \\
\midrule
\multicolumn{5}{l}{Delft--Schiedam} \\
\midrule
MAE & TabPFN & 0.278 (0.040) & $-0.0272$ ($-0.0281$, $-0.0263$) & 88.5 \\
 & SchemaGAN & 0.305 (0.034) & --- & --- \\
 & Kriging & 0.323 (0.070) & $+0.0183$ ($+0.0163$, $+0.0202$) & 43.4 \\
 & Nearest Neighbour & 0.351 (0.119) & $+0.0465$ ($+0.0429$, $+0.0500$) & 37.9 \\
\midrule
RMSE & TabPFN & 0.464 (0.058) & $-0.0328$ ($-0.0341$, $-0.0315$) & 86.6 \\
 & SchemaGAN & 0.497 (0.050) & --- & --- \\
 & Kriging & 0.504 (0.078) & $+0.0072$ ($+0.0051$, $+0.0093$) & 52.1 \\
 & Nearest Neighbour & 0.560 (0.133) & $+0.0625$ ($+0.0583$, $+0.0666$) & 31.9 \\
\midrule
\multicolumn{5}{l}{Groningen--Delfzijl} \\
\midrule
MAE & TabPFN & 0.276 (0.041) & $-0.0078$ ($-0.0086$, $-0.0069$) & 69.3 \\
 & SchemaGAN & 0.284 (0.043) & --- & --- \\
 & Kriging & 0.304 (0.046) & $+0.0201$ ($+0.0192$, $+0.0210$) & 12.4 \\
 & Nearest Neighbour & 0.326 (0.044) & $+0.0419$ ($+0.0408$, $+0.0430$) & 1.7 \\
\midrule
RMSE & TabPFN & 0.456 (0.040) & $-0.0024$ ($-0.0039$, $-0.0009$) & 60.1 \\
 & SchemaGAN & 0.459 (0.028) & --- & --- \\
 & Kriging & 0.472 (0.032) & $+0.0130$ ($+0.0118$, $+0.0142$) & 33.2 \\
 & Nearest Neighbour & 0.540 (0.055) & $+0.0816$ ($+0.0796$, $+0.0836$) & 0.8 \\
\bottomrule
\end{tabular}

\end{table}

Figure~\ref{fig:field_mae_rmse_box} shows the distributions of MAE and RMSE at held-out CPT columns in the two field sites, and Table~\ref{tab:field_summary} reports their means and paired comparisons with SchemaGAN.
Without geotechnical fine-tuning, TabPFN achieved the lowest mean MAE and RMSE among the four methods at both sites in the baseline comparison.
Depending on the site and metric, TabPFN had lower error than SchemaGAN in 60.1--88.5\% of the 2{,}000 evaluated input placements per site.
SchemaGAN had lower mean MAE and RMSE than Kriging and Nearest Neighbour at both sites.
In Delft--Schiedam section 1, however, Kriging had lower mean MAE and RMSE than SchemaGAN, and Nearest Neighbour also had lower mean MAE.
In Groningen--Delfzijl section 2, SchemaGAN's mean RMSE was lower than TabPFN's (0.459 versus 0.469).
Using predictive medians for both models (Section~\ref{sec:baselines}) preserved the site-level MAE orderings shown in Table~\ref{tab:field_summary}.
With predictive means, TabPFN retained the lowest mean RMSE at Delft--Schiedam but moved from first to third at Groningen--Delfzijl, behind SchemaGAN and Kriging.

\begin{figure*}[tbp]
  \centering
  \paperfigure{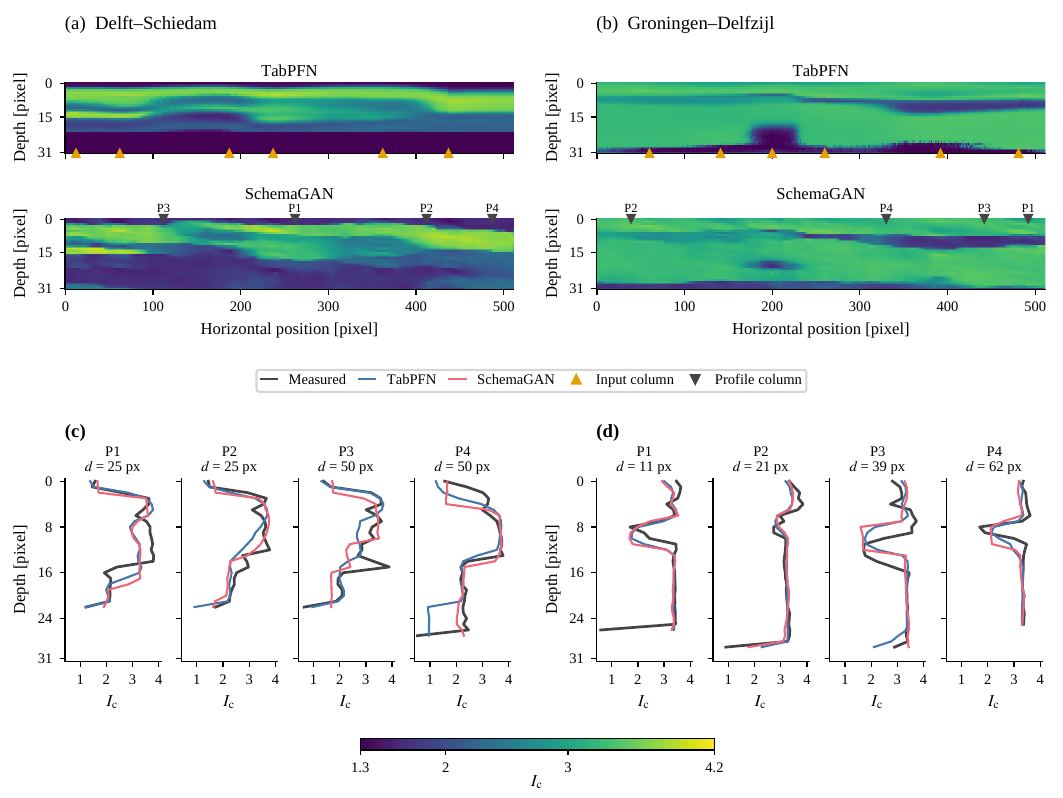}{\textwidth}{135mm}
  \caption{Field reconstructions and held-out CPT profiles of $I_\mathrm{c}$ for one section per site: Delft--Schiedam (a, c) and Groningen--Delfzijl (b, d).
    In (a) and (b), TabPFN is shown above SchemaGAN, whose predictions use inference seed 0.
    Orange triangles mark the six input columns, and grey triangles labelled P1--P4 mark the profile locations.
    Panels (c) and (d) compare predictions and measurements at four held-out columns per section.
    Profile columns were selected at distances closest to the quartiles and maximum of the horizontal distance to the nearest input column, without reference to prediction errors.
    Profile titles give this distance, $d$, in pixels.
    Axes use grid coordinates.
    Lines are drawn only at rows with CPT measurements.}
  \label{fig:field_section_profiles}
\end{figure*}

Figure~\ref{fig:field_section_profiles} shows reconstructions and held-out CPT profiles for TabPFN and SchemaGAN.
For each site, we show one input placement from the section with more held-out CPT columns.
At Delft--Schiedam, both models recovered the broad high-$I_\mathrm{c}$ layer but smoothed several measured peaks and troughs within it (Figure~\ref{fig:field_section_profiles}a,c).
Below the depths reached by the input CPTs, TabPFN produced comparatively uniform predictions that underestimated the measured values in profile~P4, whereas SchemaGAN showed greater spatial variation.
At Groningen--Delfzijl, the two reconstructions differed near the bottom of the section, while both missed much of the measured variation around rows 8--12 in profiles~P3 and~P4 (Figure~\ref{fig:field_section_profiles}b,d).
Despite its lower mean errors at both sites (Figure~\ref{fig:field_mae_rmse_box}; Table~\ref{tab:field_summary}), TabPFN was not closer to the measurements at every depth in the illustrated profiles.

\subsection{Joint reconstruction over an extended domain}
\label{sec:res_fullsite}

\begin{figure*}[tbp]
  \centering
  \paperfigure{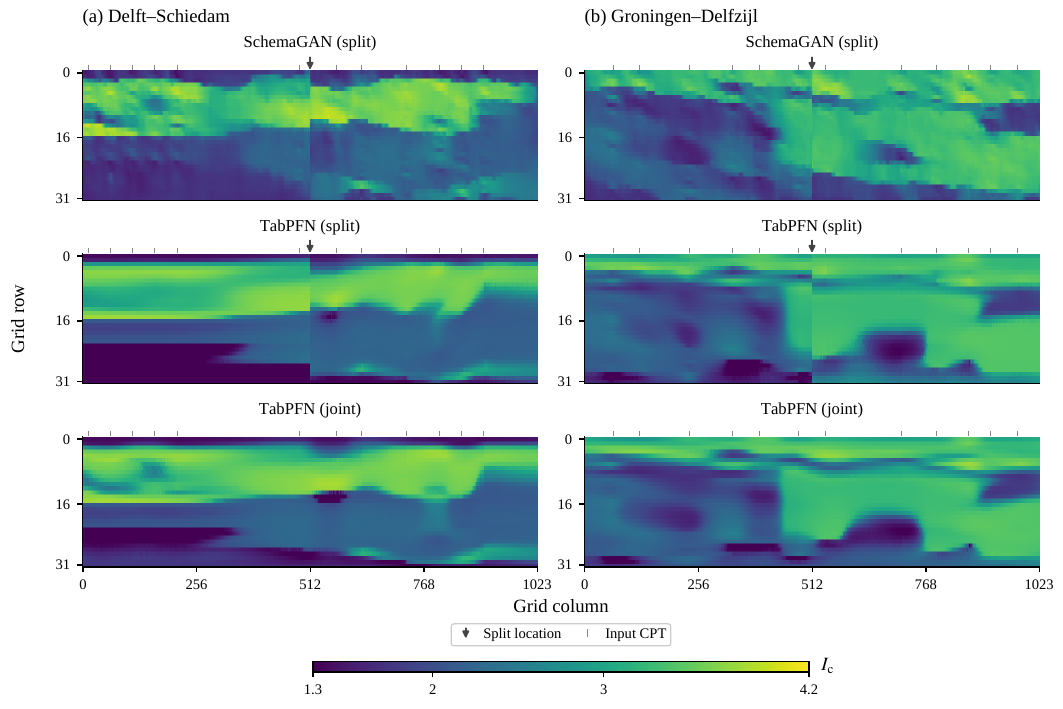}{\textwidth}{120mm}
  \caption{Reconstructions of $I_\mathrm{c}$ at (a) Delft--Schiedam and (b) Groningen--Delfzijl.
    In the split reconstructions, each section is reconstructed independently using six input CPTs; the joint TabPFN reconstruction uses the 12 input CPTs from both sections.
    TabPFN maps show predictive medians and SchemaGAN maps show individual generated reconstructions, all using inference seed 0.
    Axes use grid coordinates.}
  \label{fig:field_all_cpt_split_full}
\end{figure*}

Figure~\ref{fig:field_all_cpt_split_full} illustrates split and joint reconstructions using one of the randomly generated input placements for each section.
The joint reconstructions retained the broad low- and high-$I_\mathrm{c}$ regions visible in the split TabPFN reconstructions, with changes in local layer shapes.
At Delft--Schiedam, for example, the joint reconstruction incorporated deeper CPT observations from the right section and showed greater spatial variation below the deepest input observations in the left section (Figure~\ref{fig:field_all_cpt_split_full}a).
These reconstructions demonstrate flexibility in context size and prediction extent with fixed pretrained weights; they do not establish an improvement in prediction accuracy.
The implications for investigation updates are discussed in Section~\ref{sec:disc_incontext}.

\subsection{Effect of fine-tuning with synthetic data}
\label{sec:res_finetune}

\begin{figure*}[tbp]
  \centering
  \paperfigure{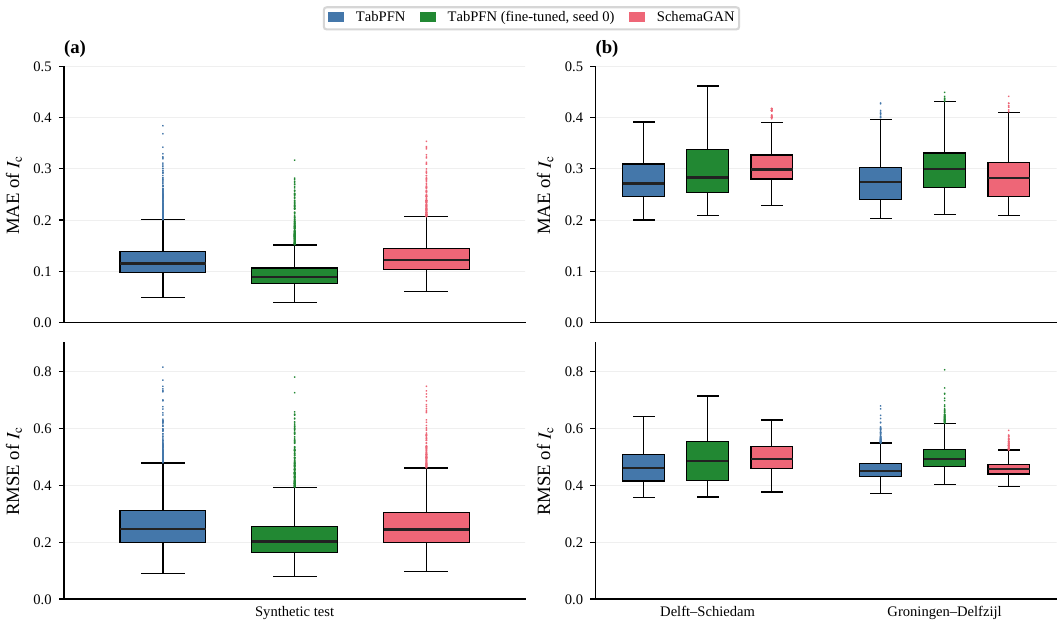}{\textwidth}{106mm}
  \caption{Distributions of MAE (top) and RMSE (bottom) in $I_\mathrm{c}$ before and after fine-tuning of TabPFN, with SchemaGAN for comparison.
    In (a), errors are computed over all pixels of the 4{,}000 synthetic test sections; in (b), errors are computed at held-out CPT columns for 2{,}000 input placements per site (1{,}000 on each of two sections).
    Whiskers extend to the most extreme values within 1.5 interquartile ranges of the box.}
  \label{fig:finetune_metric_box}
\end{figure*}

Figure~\ref{fig:finetune_metric_box} compares the error distributions before and after fine-tuning under the same input and evaluation conditions (Section~\ref{sec:experiments}).
On the synthetic test sections, fine-tuning reduced mean MAE by 22.1\% and mean RMSE by 16.0\% (Figure~\ref{fig:finetune_metric_box}a).
Both metrics showed lower medians and narrower interquartile ranges.
By contrast, mean MAE and RMSE increased by 7.2\% and 5.6\% at Delft--Schiedam and by 8.8\% and 9.7\% at Groningen--Delfzijl (Figure~\ref{fig:finetune_metric_box}b).
The mean field errors also increased in each section, and the overall trends were unchanged with a different optimisation seed (Table~\ref{tab:finetuning_seed_summary}).

\begin{figure*}[tbp]
  \centering
  \paperfigure{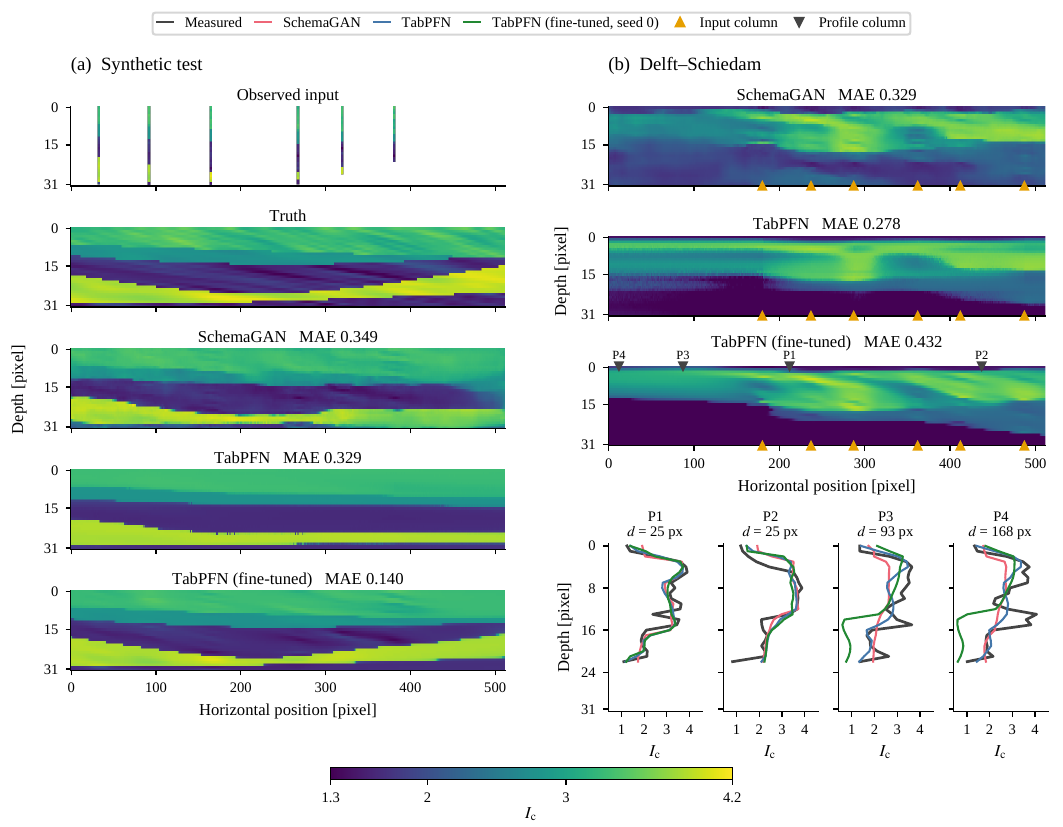}{\textwidth}{140mm}
  \caption{Examples of pronounced changes after fine-tuning, selected from the full evaluation sets: (a) a synthetic test section and (b) a field input placement at Delft--Schiedam.
    The synthetic section maximises the smaller of the MAE reductions relative to TabPFN without fine-tuning and SchemaGAN, with SchemaGAN MAE averaged over 100 inference seeds for selection.
    The field input placement maximises the MAE increase relative to TabPFN without fine-tuning.
    The displayed MAEs are calculated from the reconstructions shown, using all pixels in (a) and all held-out CPT columns in (b).
    In (b), P1--P4 mark the held-out CPT profiles selected using the distance criterion in Figure~\ref{fig:field_section_profiles}; $d$ is the distance to the nearest input column in pixels.
    Axes use grid coordinates.
    The vertical scale of the section maps and the input-column widths in (a) are exaggerated for clarity.}
  \label{fig:finetune_section_grid}
\end{figure*}

Figure~\ref{fig:finetune_section_grid} illustrates pronounced changes after fine-tuning in a synthetic section (a) and a field input placement (b).
In the synthetic example, fine-tuning improved the reconstruction of the lower high-$I_\mathrm{c}$ band, particularly its rise towards the right edge, largely beyond the rightmost input column.
Neither TabPFN without fine-tuning nor SchemaGAN reproduced this rise.
In the Delft--Schiedam example, selected for the largest MAE increase, fine-tuning lowered predicted $I_\mathrm{c}$ below the upper high-$I_\mathrm{c}$ band to the left of all input columns.
This increased the underestimation below approximately row 12 in profiles~P3 and~P4, while predictions in profiles~P1 and~P2 between input columns changed less.

\begin{figure*}[tbp]
  \centering
  \paperfigure{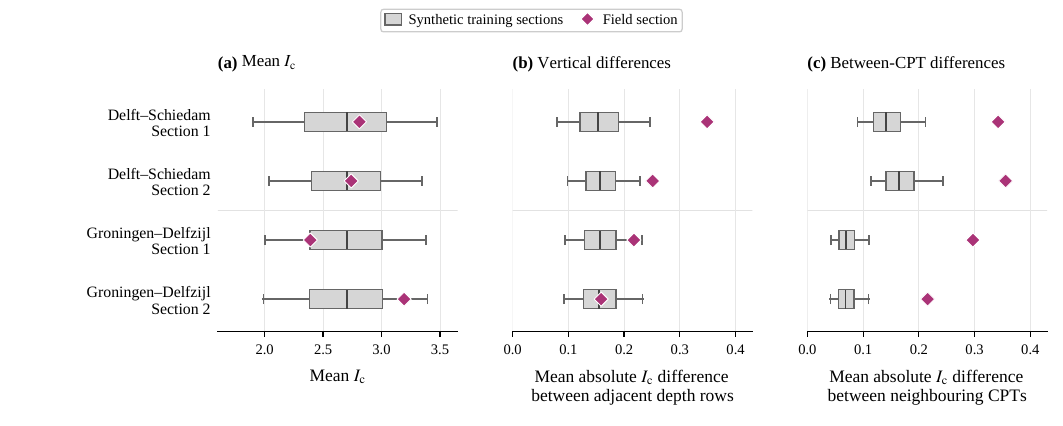}{\textwidth}{72mm}
  \caption{Comparison of $I_\mathrm{c}$ descriptors between 16{,}000 synthetic training sections and four field sections using matched CPT locations and depth rows, with the deepest non-zero row of each field CPT excluded.
    Panels show (a) mean $I_\mathrm{c}$, (b) mean absolute differences between adjacent depth rows, and (c) mean absolute differences between neighbouring CPT columns at the same depth row.
    Grey boxes span the 25th--75th percentiles, whiskers the 5th--95th percentiles, and central lines indicate medians across synthetic sections; purple diamonds indicate field values.}
  \label{fig:training_field_descriptors}
\end{figure*}

To put these contrasting fine-tuning outcomes in context, we compared the three $I_\mathrm{c}$ descriptors between the synthetic training and field sections (Figure~\ref{fig:training_field_descriptors}).
Mean $I_\mathrm{c}$ lay within the corresponding synthetic 5th--95th percentile range in all four field sections.
Vertical differences exceeded the 95th percentile in both Delft--Schiedam sections but remained within the 5th--95th percentile range in both Groningen--Delfzijl sections.
Between-CPT differences were 2.17--4.27 times the corresponding synthetic medians and exceeded the 99.9th percentile in all four field sections.
The implications of these findings for synthetic training data and field validation are discussed in Section~\ref{sec:disc_finetuning}.

\section{Discussion}
\label{sec:discussion}

\subsection{Transfer from general tabular pretraining}
\label{sec:disc_transfer}

Building on applications of TabPFN to soil-property prediction and digital soil mapping~\cite{saito2026sitecharacterization,schmidinger2026kriging}, we evaluated its ability to reconstruct two-dimensional $I_\mathrm{c}$ sections from sparse CPT observations under varying observation conditions, using grid coordinates as the only features.
Our results provide initial evidence that general tabular pretraining can support this reconstruction without geotechnical fine-tuning.
With its pretrained weights held fixed, TabPFN recovered the main layer geometry in a representative synthetic section and achieved mean errors close to those of SchemaGAN across the synthetic test sections (Section~\ref{sec:res_synthetic}).
In the baseline field comparison, it achieved the lowest site-level mean MAE and RMSE among the four methods at both sites (Section~\ref{sec:res_field}).

Both Kriging and TabPFN combine prior assumptions with the observed data: Kriging uses a specified covariance model, whereas TabPFN draws on its general tabular prior.
This prior covers diverse relationships among variables, including spatial relationships~\cite{grinsztajn2026tabpfn3}.
Subsurface patterns may share statistical regularities with the synthetic pretraining data, allowing TabPFN to use learned relationships to represent local smoothness, larger-scale trends and interactions between the coordinate variables.
However, we did not analyse the model's internal representations or isolate the contributions of these learned relationships.
The reconstructions therefore do not establish that the model has geological understanding or recovers the processes that formed the ground.
Future work could investigate the internal representations underlying these spatial predictions through embedding analyses, as explored in previous geotechnical applications of TabPFN~\cite{saito2026interpretable}.

TabPFN and SchemaGAN differ in their pretraining data, input representations and inference procedures, so their relative errors do not isolate the effect of either the prior or the network architecture (Section~\ref{sec:baselines}).
SchemaGAN's field errors may also depend on how well its training sections represent site layering~\cite{camposmontero2025schemagan}.
On the synthetic sections, the crossing of the distance curves (Figure~\ref{fig:observation_conditions}b) shows that the MAE ordering of TabPFN and SchemaGAN also depends on proximity to observations.
TabPFN achieved lower MAE than SchemaGAN farther from observations while producing smaller spatial differences overall than those in the target sections (Table~\ref{tab:spatial_difference_ratios}).
TabPFN's preservation of layer continuity and SchemaGAN's recovery of a thin inclined layer in different examples illustrate why local layer geometry should be assessed alongside mean errors (Figure~\ref{fig:synthetic_section_grid}b,c).
Overall, these results suggest that, without geotechnical fine-tuning, TabPFN can provide initial section reconstructions for examining the main layer geometry between sparse CPTs.

\subsection{In-context inference and investigation updates}
\label{sec:disc_incontext}

With in-context inference, we can incorporate additional CPT observations and extend the prediction domain without updating TabPFN's weights.
In the synthetic experiments, TabPFN's mean MAE decreased at every tested increase in CPT count from 3 to 48, indicating that it benefited from additional observations supplied as context (Section~\ref{sec:res_conditions}).
SchemaGAN's response to additional CPTs should be interpreted in relation to its training configuration: its released training implementation uses six input columns per section.
The released model accepted other CPT counts in our experiments, but its mean MAE increased above six CPTs, possibly reflecting changes in observation density and column spacing relative to this configuration.
Retraining SchemaGAN on larger datasets with more varied layer geometries and mixed CPT counts could change this response and the comparison with TabPFN.
For larger prediction domains, SchemaGAN can combine predictions from overlapping sections~\cite{camposmontero2025digitaltwin}.
TabPFN instead combined the input CPT observations from each site's two sections into a single context, allowing these observations to inform predictions throughout the joint grid (Section~\ref{sec:res_fullsite}).

These results suggest that a model pretrained on general tabular data can provide an initial section reconstruction with default preprocessing, without requiring a representative geotechnical training dataset.
During sequential site investigation, CPT observations from each stage could then be added to the context to update predictions and their uncertainty in previously reconstructed areas and extend the reconstruction as the investigated area grows.
The synthetic prediction-interval results (Section~\ref{sec:res_intervals}) motivate testing whether TabPFN's predictive uncertainty can inform the selection of subsequent CPT locations, once its intervals have been validated at held-out field CPTs.
Zuada Coelho et al.~\cite{zuadacoelho2026investigations} used deep reinforcement learning to select investigation locations adaptively in synthetic geotechnical sections.
Future work could couple a reinforcement learning policy with TabPFN's prediction updates.
Such a policy would require its own training and validation, balancing investigation cost against the value of additional observations for site characterisation and engineering decisions.

Beyond adding CPT observations, in-context inference allows the same pretrained model to be used with different features and targets, so the coordinate-only formulation could serve as a baseline for evaluating richer inputs.
Possible extensions include topography, other CPT-derived parameters, and information from borehole logs, geophysical measurements or geological interpretations.
Implementing these extensions would require aligning depth references, accounting for the spatial extent represented by each measurement, defining compatible prediction targets, and specifying which features are available at query locations.

\subsection{Fine-tuning and geotechnical training data}
\label{sec:disc_finetuning}

Under the training and checkpoint-selection procedure used here, fine-tuning improved synthetic test accuracy but increased mean MAE and RMSE at both field sites (Section~\ref{sec:res_finetune}).
In the Delft--Schiedam example, fine-tuning increased underestimation to the left of all input CPTs while predictions between them changed less (Figure~\ref{fig:finetune_section_grid}b).
In the descriptor comparison, between-CPT differences in all four field sections exceeded the 99.9th percentile of the corresponding synthetic training distributions, while vertical differences exceeded the synthetic 95th percentile only in the Delft--Schiedam sections (Figure~\ref{fig:training_field_descriptors}).
One possible explanation for the field deterioration is that fine-tuning adapted the model more closely to the layer geometries and within-layer variability represented in the synthetic training data, reducing its ability to generalise to field conditions with different characteristics.
We compared descriptors calculated on the section grids, so the observed differences may reflect how the field CPT data were preprocessed and averaged into grid rows (Section~\ref{sec:data}).
The present experiments do not establish the cause of the field deterioration or rule out overfitting to individual training sections.
We did not separately assess the effects of the training distribution, optimisation hyperparameters and checkpoint selection.

Future comparisons across different ground conditions and synthetic data-generation procedures should investigate the causes of the field deterioration and examine whether site information is better introduced through context or weight updates.
Training sets could vary layer geometries, within-layer variability and observation patterns, with generation parameters informed by field CPTs.
Context selection could use observations from similar sites, drawing on the concept of site recognition in geotechnical engineering~\cite{phoon2022challenges}.
Site characteristics, such as depositional history or stratigraphic structure, could guide the selection of context observations or synthetic training sections, and could also be included as features when $\mathcal C$ combines observations from multiple sites.
Comparisons of context selection and fine-tuning at sites not used to design these approaches, using the same available field information, would test whether each approach improves field prediction and which is preferable when site information is limited.

\subsection{Limitations and future validation}
\label{sec:disc_limitations}

Across the field accuracy experiments, we evaluated two fixed sections at each of two sites in the Netherlands, using random input placements to assess sensitivity to observation selection.
These placements do not provide independent tests across geological settings, and accuracy at locations without CPT measurements remains unverified.
We have also not tested how field accuracy changes with CPT count or how TabPFN's predictions vary with its inference seed.
Further validation should include independent sites with different depositional histories, stratigraphic structures and investigation patterns.
Regarding spatial representation, the field targets are depth-averaged $I_\mathrm{c}$ values, so the results do not establish reconstruction of variations finer than the grid intervals.
Grid indices also omit physical spacings and aspect ratios (Section~\ref{sec:tabpfn}).
For the joint reconstructions, we placed the two section grids at each site side by side without rescaling and assumed that equal row indices represented the same depth interval (Section~\ref{sec:experiments}).
The illustrated comparisons cover one input placement per site; they do not establish an accuracy benefit of the combined context across input placements or assess sensitivity to the assumed depth correspondence (Section~\ref{sec:res_fullsite}).
More generally, sensitivity to grid resolution and the resulting changes in observation aggregation remains untested.
Further work should examine how these factors and site geometry affect reconstruction accuracy, and how coordinate representations and internal preprocessing influence TabPFN's representation of spatial anisotropy.
For the difference ratios (Table~\ref{tab:spatial_difference_ratios}), we evaluated TabPFN's pixelwise predictive medians and SchemaGAN's individual generated reconstructions; using the pixelwise median of SchemaGAN's 100 reconstructions instead changed the across-section medians of both ratio types by less than 0.03 at every tested separation and preserved the orderings of these ratios reported in Section~\ref{sec:res_synthetic}.
Neither the difference ratios, which do not distinguish within-layer changes from layer transitions, nor MAE and RMSE directly measure boundary locations or layer connectivity.
Dedicated evaluation of stratigraphic geometry and engineering-property assignments is therefore needed to assess the engineering usefulness of the reconstructions.

We evaluated prediction intervals only on synthetic sections with six input CPTs; field point-prediction accuracy does not establish interval calibration at the field sites.
Although TabPFN's mean pointwise coverage was close to nominal levels on these sections, this average does not establish calibration at every location or simultaneous coverage of an entire section.
SchemaGAN's undercoverage concerns the narrow intervals obtained here from Monte Carlo dropout with a pixelwise normal approximation; we did not evaluate other uncertainty formulations.
These narrow intervals are consistent with the limited output variation under dropout reported for pix2pix~\cite{isola2017pix2pix}, on which SchemaGAN is based.
Future work should test interval calibration at other CPT counts and at held-out field CPTs, including at independent sites.
To support engineering decisions, future studies should also develop joint representations of uncertainty that account for spatial dependence.
These representations should be used to propagate uncertainty in reconstructed ground conditions through geotechnical analyses, including uncertainty in any conversion from $I_\mathrm{c}$ to engineering properties.
Case studies of sequential investigation (Section~\ref{sec:disc_incontext}) could then evaluate whether additional observations improve predictions of engineering responses and decisions about safety, serviceability and cost.
A longer-term objective is to generate spatially coherent alternative ground models that represent uncertainty in layer continuity, boundary geometry and material distribution.

\section{Conclusion}
\label{sec:conclusion}

This study provides initial evidence that TabPFN, pretrained on general synthetic tabular data, can reconstruct two-dimensional $I_\mathrm{c}$ sections from grid coordinates and sparse CPT observations without geotechnical fine-tuning.
In the baseline comparison, TabPFN had mean prediction errors close to those of SchemaGAN on synthetic sections and the lowest site-level mean MAE and RMSE among the four methods at held-out CPT locations at both field sites.
With fixed pretrained weights, TabPFN's mean MAE decreased at each tested increase in CPT count on the synthetic sections, and the model reconstructed the two field-section grids at each site jointly from their combined input CPTs.
Without additional calibration, its pointwise prediction intervals achieved mean coverage close to the nominal levels on synthetic sections with six input CPTs, with narrower mean widths than those of the Kriging implementation compared here.
Under the training and checkpoint-selection procedure used here, fine-tuning on synthetic sections improved synthetic test accuracy but increased mean MAE and RMSE at both field sites.

Future research should investigate how similarity in ground conditions can guide the selection of additional context observations from other sites, and whether synthetic training data informed by field CPTs can improve field prediction after fine-tuning.
Validation at independent field sites and in sequential investigations should assess prediction accuracy and interval calibration across different ground and observation conditions.
Further studies should account for the spatial dependence of prediction errors when propagating uncertainty through geotechnical analyses to assess the implications for investigation and design decisions.
More broadly, the results from adding CPT observations and extending the prediction domain with fixed pretrained TabPFN weights indicate that general-purpose tabular foundation models offer a promising route for updating subsurface predictions from sparse and evolving site-investigation data.

\section*{Data availability}

The synthetic dataset is available at \url{https://doi.org/10.5281/zenodo.13143431}.
Code for reproducing the synthetic experiments is available at \url{https://github.com/tigtie09/tabpfn-for-subsurface-schematisation}.

\clearpage
\appendix
\setcounter{table}{0}

\section{Inference and fine-tuning settings}
\label{app:finetuning}

TabPFN inference (float32 precision) and both fine-tuning runs used PyTorch 2.11.0 with CUDA 12.8, and SchemaGAN inference used TensorFlow 2.15.1.
Each fine-tuning step used one section and 2{,}048 query locations sampled uniformly without replacement from its unobserved pixels.
Training used one ensemble member with a softmax temperature of 0.9, cycling through the four preprocessing combinations in Section~\ref{sec:tabpfn}.
Preprocessing was fitted only to the context observations; context and query targets were standardised using the mean and standard deviation of the context targets before the selected target transformation.
We used AdamW with a peak learning rate of $10^{-5}$ and weight decay of 0.01, with 500 steps of linear warmup followed by cosine decay.
The gradient norm was clipped at 1.0, and automatic mixed precision was used.

The training and validation inputs were drawn once per section using the baseline sampling procedure (Section~\ref{sec:experiments}).
Because each split's inputs were sampled with seed 20232023 in lexicographic file-path order, placements coincided at corresponding positions in the validation and test splits and the first 4{,}000 training sections, although the target sections differed.
We therefore also evaluated TabPFN before and after fine-tuning on the same 4{,}000 synthetic test sections with fresh six-column inputs drawn by the same procedure (seed 20261022), none of which repeated a training, validation or baseline test placement.
For optimisation seed 0, fine-tuning reduced mean MAE and RMSE by 22.6\% and 16.5\% with fresh inputs, compared with 22.1\% and 16.0\% with the baseline inputs.
To keep the evaluation pixels fixed across input placements, we also scored only pixels unobserved under both placements; the improvements from fine-tuning persisted.

We repeated fine-tuning with optimisation seed 1, keeping the training and validation data, input placements, initial weights and hyperparameters fixed (Table~\ref{tab:finetuning_seed_summary}).
Both runs were validated after each epoch using the inference settings in Section~\ref{sec:tabpfn}.
The mean validation MAE in $I_\mathrm{c}$ over unobserved pixels decreased at every epoch, from 0.1271 before fine-tuning to 0.0981 and 0.0983 for seeds 0 and 1, respectively; both runs selected epoch 5.
For seed 1, mean training CRPS in the internal target space decreased from 0.1482 at epoch 1 to 0.1243 at epoch 5.
Mean training MSE decreased from 0.3180 to 0.2298 by epoch 4 before increasing to 0.3396 at epoch 5.
The seed 1 run, including validation, took 18.1 hours on one NVIDIA RTX 6000 Ada Generation GPU and one CPU thread, with a peak allocated GPU memory of 3.21 GiB.

\begin{table}[tbp]
  \centering
  \caption{Mean MAE and RMSE in $I_\mathrm{c}$ before and after fine-tuning with optimisation seeds 0 and 1, using the baseline input placements.
    Synthetic errors are averaged over the 4{,}000 test sections using all pixels; field errors are calculated at held-out CPT columns and averaged over 2{,}000 input placements per site.}
  \label{tab:finetuning_seed_summary}
\begin{tabular}{llrrr}
\toprule
Dataset & Metric & Before fine-tuning & Seed 0 & Seed 1 \\
\midrule
Synthetic & MAE & 0.1228 & 0.0956 & 0.0957 \\
 & RMSE & 0.2638 & 0.2215 & 0.2217 \\
\midrule
Delft--Schiedam & MAE & 0.2777 & 0.2978 & 0.3013 \\
 & RMSE & 0.4644 & 0.4905 & 0.4943 \\
\midrule
Groningen--Delfzijl & MAE & 0.2760 & 0.3002 & 0.2957 \\
 & RMSE & 0.4565 & 0.5006 & 0.4934 \\
\bottomrule
\end{tabular}

\end{table}

\FloatBarrier
\Needspace{0.6\textheight}
\section{Spatial variation in synthetic reconstructions}
\label{app:spatial_variation}
\setcounter{table}{0}

\begin{table}[!htbp]
  \centering
  \caption{Reconstructed-to-target ratios of absolute $I_\mathrm{c}$ differences for the 4{,}000 synthetic test sections using baseline inputs.
    Cells report medians across sections; SchemaGAN ratios are first averaged over 100 individually evaluated reconstructions within each section.
    The two blocks use mean and median absolute differences over all pixel pairs at each indicated grid separation, including pairs involving input pixels.
    For Nearest Neighbour at the tested horizontal separations from 1 to 32 columns, more than half of the pixel pairs have identical values in every section, giving zero ratios of median absolute differences.
    Ratios below one indicate smaller differences than in the target sections; neither type of ratio isolates within-layer variation or measures boundary locations or layer connectivity.}
  \label{tab:spatial_difference_ratios}
\begingroup
\fontsize{10}{12}\selectfont
\setlength{\tabcolsep}{2.5pt}
\begin{tabular*}{\textwidth}{@{\extracolsep{\fill}}l*{12}{r}@{}}
\toprule
Method & \multicolumn{4}{c}{\shortstack{Vertical separation\\(rows)}} & \multicolumn{8}{c}{\shortstack{Horizontal separation\\(columns)}} \\
\cmidrule(lr){2-5}\cmidrule(lr){6-13}
 & 1 & 2 & 4 & 8 & 1 & 2 & 4 & 8 & 16 & 32 & 64 & 128 \\
\midrule
\multicolumn{13}{l}{Ratios of mean absolute differences} \\
TabPFN & 0.65 & 0.75 & 0.86 & 0.94 & 0.82 & 0.53 & 0.40 & 0.37 & 0.40 & 0.48 & 0.62 & 0.76 \\
SchemaGAN & 0.76 & 0.83 & 0.91 & 0.97 & 1.23 & 1.03 & 0.86 & 0.77 & 0.74 & 0.77 & 0.83 & 0.88 \\
Kriging & 0.76 & 0.85 & 0.91 & 0.91 & 0.30 & 0.31 & 0.31 & 0.32 & 0.37 & 0.47 & 0.65 & 0.83 \\
Nearest Neighbour & 0.85 & 0.87 & 0.89 & 0.93 & 0.38 & 0.38 & 0.39 & 0.41 & 0.47 & 0.61 & 0.88 & 1.03 \\
\midrule
\multicolumn{13}{l}{Ratios of median absolute differences} \\
TabPFN & 0.34 & 0.41 & 0.56 & 0.93 & 0.46 & 0.29 & 0.21 & 0.18 & 0.19 & 0.23 & 0.34 & 0.51 \\
SchemaGAN & 0.59 & 0.64 & 0.75 & 0.94 & 1.22 & 1.18 & 0.94 & 0.81 & 0.73 & 0.72 & 0.77 & 0.82 \\
Kriging & 1.12 & 1.20 & 1.27 & 0.92 & 0.36 & 0.36 & 0.36 & 0.37 & 0.40 & 0.52 & 0.79 & 1.12 \\
Nearest Neighbour & 0.78 & 0.80 & 0.81 & 0.88 & 0.00 & 0.00 & 0.00 & 0.00 & 0.00 & 0.00 & 0.58 & 1.03 \\
\bottomrule
\end{tabular*}
\endgroup

\end{table}

\FloatBarrier
\bibliographystyle{elsarticle-num}
\bibliography{references}

\end{document}